\def\year{2022}\relax
\documentclass[letterpaper]{article} 
\usepackage{aaai22}  
\usepackage{times}  
\usepackage{helvet}  
\usepackage{courier}  
\usepackage[hyphens]{url}  
\usepackage{graphicx} 
\usepackage{natbib}  
\usepackage{caption} 
\DeclareCaptionStyle{ruled}{labelfont=normalfont,labelsep=colon,strut=off} 
\usepackage{algorithm}
\usepackage{algorithmic}
\usepackage{listings}
\usepackage{amsmath,amssymb}
\usepackage{subcaption}	

\usepackage{newfloat}
\usepackage{listings}
\floatstyle{ruled}
\newfloat{listing}{tb}{lst}{}
\floatname{listing}{Listing}
\usepackage{xcolor}
\definecolor{codegreen}{rgb}{0,0.6,0}
\definecolor{codegray}{rgb}{0.5,0.5,0.5}
\definecolor{codepurple}{rgb}{0.58,0,0.82}
\definecolor{backcolour}{rgb}{0.97,0.97,0.97}

\lstdefinestyle{mystyle}{
    backgroundcolor=\color{backcolour},   
    commentstyle=\color{codegreen},
    keywordstyle=\color{blue},
    numberstyle=\tiny\color{codegray},
    stringstyle=\color{codepurple},
    basicstyle=\ttfamily\scriptsize,
    captionpos=b,               
}

\newcommand{\Sectionref}[1]{Section~\ref{#1}}
\newcommand{\sectionref}[1]{section~\ref{#1}}

\title{Tools and Practices for Responsible AI Engineering}
\author {
    Ryan Soklaski$^*$,
    Justin Goodwin$^*$,
    Olivia Brown,
    Michael Yee,
    and Jason Matterer
}
\affiliations {
    MIT Lincoln Laboratory\\
    \{ryan.soklaski, jgoodwin, olivia.brown, myee, jason.matterer\}@ll.mit.edu\\
    \textsuperscript{\rm $^*$} These authors contributed equally to this work \\
}

\begin{document}

\maketitle

\makeatletter{\renewcommand*{\@makefnmark}{}

\footnotetext{\\ DISTRIBUTION STATEMENT A. Approved for public release. Distribution is unlimited. (See Acknowledgements for details.)}\makeatother}

\begin{abstract}

Responsible Artificial Intelligence (AI)---the practice of developing, evaluating, and maintaining accurate AI systems that also exhibit essential properties such as robustness and explainability---represents a multifaceted challenge that often stretches standard machine learning tooling, frameworks, and testing methods beyond their limits.
In this paper, we present two new software libraries---\texttt{hydra-zen} and the \texttt{rAI-toolbox}---that address critical needs for responsible AI engineering. \texttt{hydra-zen} dramatically simplifies the process of making complex AI applications configurable, and their behaviors reproducible. The \texttt{rAI-toolbox} is designed to enable methods for evaluating and enhancing the robustness of AI-models in a way that is scalable and that composes naturally with other popular ML frameworks. We describe the design principles and methodologies that make these tools effective, including the use of property-based testing to bolster the reliability of the tools themselves. Finally, we demonstrate the composability and flexibility of the tools by showing how various use cases from adversarial robustness and explainable AI can be concisely implemented with familiar APIs.

\end{abstract}

\section{Introduction}

Responsible\footnote{Also commonly referred to as ``trustworthy'' or ``ethical''} Artificial Intelligence (AI) is the practice of creating and maintaining AI systems that are not only accurate, but also exhibit important qualities such as: robustness, safety, security, and privacy; fairness and inclusiveness; transparency, interpretability, and explainability; accountability and governability; and promotion of societal and environmental well-being. There is growing recognition that more attention needs to be given to these critical principles, as evidenced by the publication of more than 400 policy documents \cite{shneiderman2021responsible}, including guidance from leading industry and government organizations \cite{, microsoft2021responsible, googleResponsible, ibmTrustworthy, dod2020ethical, european2019ethics}. However, doing work in the space of responsible AI puts researchers and engineers at the edge of what is currently capable with existing tools.

\begin{figure}[t]
\includegraphics[scale=0.55]{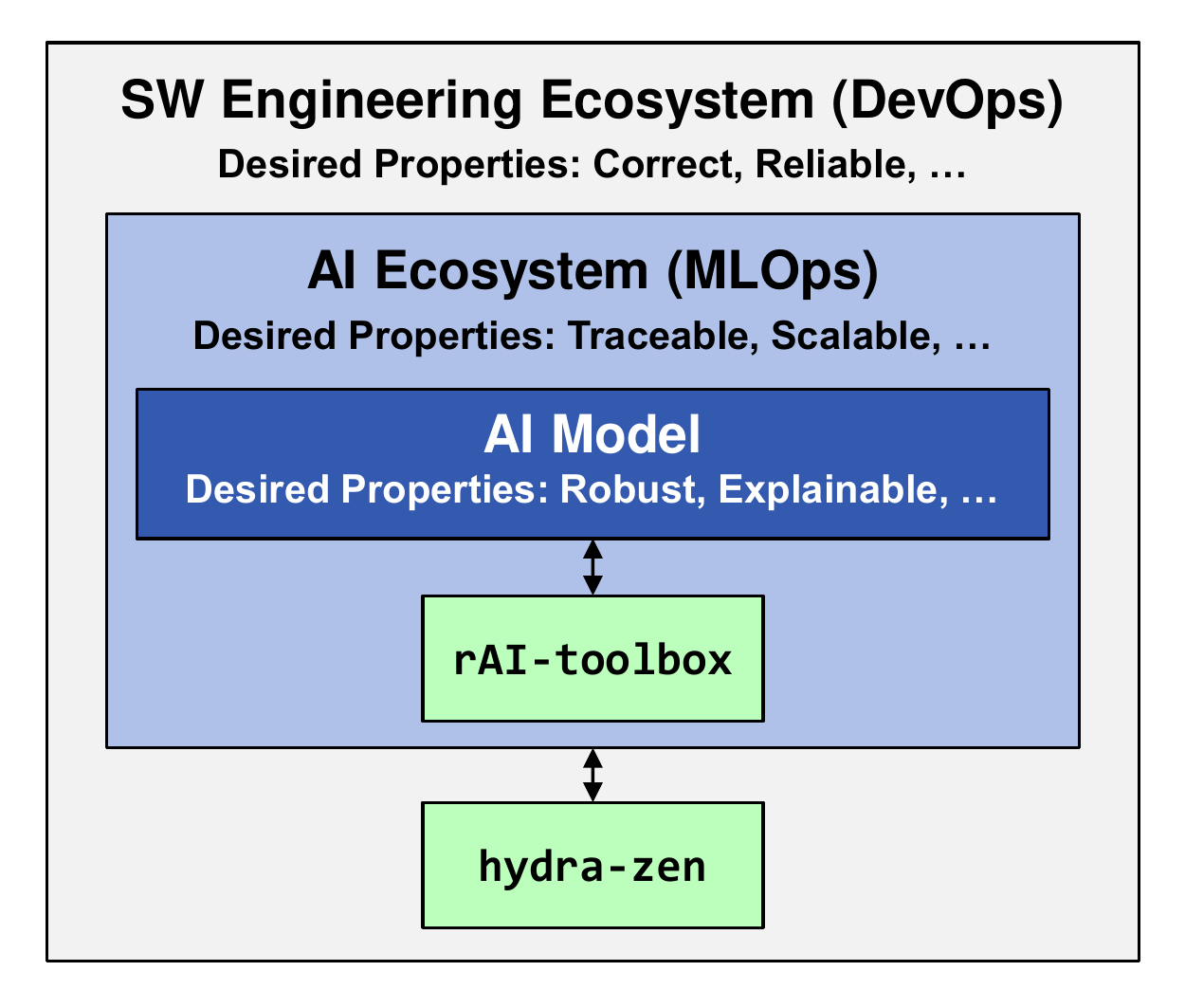}
\centering
\caption{\textbf{Framework for Responsible AI Engineering}, with new tools introduced in this paper shown in green. \texttt{hydra-zen} simplifies the process of making complex software (SW) applications configurable and reproducible with Hydra, thereby increasing the traceability and scalability of ML workflows. The \texttt{rAI-toolbox} provides first-class support for advanced ML techniques (such as robust training) that go beyond the standard train/test/deploy paradigm supported by popular high-level ML APIs.}
\label{fig:responsible_ai}
\end{figure}

Engineering reliable software systems is already a challenging endeavor, although modern DevOps (and DevSecOps) practices and tools help manage risk. Machine learning (ML) applications have the added complexity of being data driven, and thus require additional support for careful tracking, management, and monitoring of datasets, experiments, and models across the machine learning system lifecycle (MLOps). Additional properties required by responsible AI systems, such as robustness and explainability, are active research areas with a developing ecosystem of tools.


In this paper, we present two software libraries\footnote{\texttt{hydra-zen} is available, and the \texttt{rAI-toolbox} will be open-sourced soon, here: https://github.com/mit-ll-responsible-ai/} that address critical needs for responsible AI engineering (see Figure~\ref{fig:responsible_ai}). \texttt{hydra-zen} extends the popular Hydra framework~\cite{Yadan2019Hydra} to standardize the process of making complex AI applications configurable and their behaviors reproducible. It provides specialized tools that dramatically simplify the process of adopting the Hydra framework and that eliminate major sources of technical debt that this framework typically creates. The \texttt{rAI-toolbox} is designed to enable methods for evaluating and enhancing the robustness of AI models in a way that is scalable and that composes naturally with other popular ML frameworks, such as PyTorch~\cite{NEURIPS2019_9015}.

We demonstrate the composability and flexibility of the tools by showing how two different use cases from adversarial robustness and explainable AI can be concisely implemented with familiar APIs. We then conclude with a final example that computes robustness curves for two different models using both \texttt{hydra-zen} and the \texttt{rAI-toolbox}, to show how robustness analyses can be easily scaled. Both libraries rely on property-based testing~\cite{PBT} and automatic test case generation, leveraging the Hypothesis library~\cite{maciver2019hypothesis,maciver2020test}, in order to ensure that the tools used for evaluating AI systems are themselves reliable.

\section{Configurable and Reproducible AI Systems via Hydra and hydra-zen} \label{section:hydra-zen}

Conducting experiments and analyses in a reproducible manner is a critical facet of responsible AI engineering practices. The need to fastidiously document and organize one's work in a systematic way is of manifest importance to any scientific endeavor. That being said, AI systems are often sprawling software enterprises whose pillars---data processing pipelines, model architectures, and training \& testing frameworks---are each complex, hierarchical systems whose states are challenging to record. While some popular tools, such as Weights \& Biases~\cite{wandb}, Comet~\cite{CometML}, and MLFlow~\cite{MLFlow}, provide support for tracking experiments in the name of reproducibility, they ultimately provide only ``shallow," incomplete summaries of AI systems. By contrast, the Hydra framework~\cite{Yadan2019Hydra} has designed a rich, domain-specific language for describing configurations of complex software applications; this framework, leveraged via our user-facing \texttt{hydra-zen} Python library~\cite{soklaski2021hydrazen}, enables AI systems to be made thoroughly configurable and reproducible. 

In general, Hydra and \texttt{hydra-zen} standardize the process of designing AI systems that are:

\begin{description}
\item[Configurable:] All aspects of one's Hydra-based application---including deeply nested components---can be configured from a single interface.
\item[Reproducible:] Each run of the application is self-documenting; the full configuration of the software is saved alongside the results of that run.
\item[Scalable:] Multiple configurations of the application can be launched---to sweep or search over configuration subspaces---using a variety of local and distributed job-launcher methods.
\end{description}

\begin{lstlisting}[frame=single,%numbers=none,numberstyle=\tiny, numbersep=0pt,
float=tp,
numbers=none,
xleftmargin=4pt,
language=Python,label={lst:hzen-example},
caption={Using \texttt{hydra-zen} to automatically generate a configuration for a popular optimizer from the PyTorch framework. This ``config" can be serialized to a YAML file by Hydra in order to record its exact configuration for purposes of reproducibility.}]
>>> from hydra_zen import builds, to_yaml
>>> from torch.optim import Adam

>>> OptimConfig = builds(Adam, ...)
>>> print(to_yaml(OptimConfig))
_target_: torch.optim.adam.Adam
params: ???
lr: 0.001
betas:
- 0.9
- 0.999
eps: 1.0e-08
weight_decay: 0
amsgrad: false
\end{lstlisting}

At the core of the Hydra framework is a YAML-based, domain-specific language that enables the description and configuration of hierarchical structures in a Python-based software application. Thus, even nested components of one's AI system can be configured from a single interface, and a complete configuration of the entire system can be represented via a human-readable YAML file. Additionally, one can identify so-called ``configuration groups" in the application's structure, so that entire sections of the system---e.g., a particular data pre-processing procedure---can be ``swapped" en-masse in an ergonomic way. Reproducibility is a natural consequence of this configurability: each job launched by Hydra is documented by---and can be fully replicated by---the YAML configuration that is automatically recorded for that job.

\texttt{hydra-zen} provides Hydra users with elegant tools for automatically generating and customizing Hydra-compatible configurations. Without these tools, users are faced with hand-writing and maintaining YAML configurations (or their Python-based equivalents) for their entire software application; such a process is manually intensive, error-prone, repetitive, and is a major source of technical debt \cite{DRY}. \texttt{hydra-zen} eliminates these costs by enabling a Python-centric, ergonomic workflow for dynamically populating and automatically validating configurations for one's entire software application. For example, Listing \ref{lst:hzen-example} demonstrates the use of \texttt{hydra-zen}'s \texttt{builds} function to dynamically generate a configuration for an optimizer from the PyTorch library, based on that optimizer's default values. Thus, \texttt{hydra-zen} makes it tractable for both large-scale AI systems and smaller, rapid-prototype systems to be designed to be configurable, reproducible, and scalable via the Hydra framework.

\texttt{hydra-zen}'s configuration-creation tools also enable a unique, configuration-based functionality modification framework, which can be used to augment the behavior of the various components of a system without modifying the system's source code. For example, rich runtime type-checks of configured values, and schema-based data validation checks, can be ``injected" into the system's various interfaces---and be subsequently enabled or disabled---by solely modifying the configuration of the system. The ability to effectively patch the behaviors of a system via its configuration is highly valuable from a system-maintenance perspective; this capability is not native to Hydra, but is a ``higher-order" capability that is enabled by the design of the functions provided by \texttt{hydra-zen}.      

Ultimately, Hydra and \texttt{hydra-zen} are uniquely well-suited to enable rigorous reproducibility in AI systems, by-design.


\section{Responsible AI Toolbox}

While \texttt{hydra-zen} contributes to responsible AI practices through enabling the ecosystem surrounding AI model development and deployment to be more transparent, there is also a need for the AI models themselves to be engineered to obey the set of responsible AI principles. There is a growing effort to innovate methods for assessing and bolstering the robustness and explainability of AI systems---two key facets of responsible AI. While a variety of tools exist that implement common techniques from the research community, such as Foolbox~\cite{rauber2017foolbox}, CleverHans~\cite{papernot2018cleverhans}, and IBM's Adversarial Robustness Toolbox and AI Explainability 360~\cite{nicolae2018adversarial,aix360-sept-2019}, these tools have yet to be fully integrated into existing MLOps frameworks. Many of the existing tools are designed to provide a large library of techniques for evaluating or enhancing robustness in a framework agnostic API; this has a handful of limitations: (1) support for recent techniques is accomplished by relying heavily on research implementations, which can result in brittle and specialized code, being domain or even dataset-specific and (2) support for multiple frameworks results in un-scalable code that is difficult to build from. These observations inspired a handful of design principles for the \texttt{rAI-toolbox} that we explain in the following sections. 

In \sectionref{ss:core_problems}, we identify key low-level problems that that often need to be solved for robust AI tasks. Solutions to these problems form the core of the \texttt{rAI-toolbox}. In \sectionref{ss:composability}, we explain the advantages of designing the \texttt{rAI-toolbox} around a single deep learning framework. \Sectionref{ss:scalability} describes how the \texttt{rAI-toolbox} supports scalable workflows by being compatible with existing tools that enable distributed computation. Finally, in \sectionref{ss:reliability}, we describe how we employ property-based testing (PBT) to ensure that the \texttt{rAI-toolbox} itself is reliable.

\subsection{Core Problems in Robust AI}\label{ss:core_problems}
Whereas standard model-training frameworks are designed to refine the parameters of the machine learning \emph{model} (i.e., architecture and weights), methods for studying the robustness (and often other responsible AI properties) of the model naturally involve analyses of and optimizations over the \emph{data} (i.e., inputs to the model and representations extracted by the model). This optimization over the data space increases the complexity of the responsible AI engineering workflow over that of the standard setting.

For example, consider the standard optimization objective for training a model, $f_\theta$, parameterized by $\theta$, where $x$ and $y$ represent the data input and corresponding output sampled from a data distribution, $D$, and $\mathcal{L}$ is the loss function to be minimized:
\begin{equation} \label{eq:standard}
    \min\limits_{\theta \in \Theta} \mathbb{E}_{(x,y)\sim D} [\mathcal{L}(f_\theta(x),y)].
\end{equation}
Note that here, the data samples are fixed, and the search is done over the model's parameter space. Once the model is trained, its performance is often evaluated using an independent (but usually fixed) set of data samples from the same data distribution.

Now, consider the process of solving for a worst-case (i.e., ``adversarial") perturbation to a data input to fool the model into producing an incorrect output, which is common practice when assessing the adversarial robustness of the model \cite{carlini2019evaluating}. The perturbation, $\delta$, is optimized to maximize loss against the true output, $y$, subject to a constraint set, $\Delta$:
\begin{equation} \label{eq:advpert}
    \max\limits_{\delta \in \Delta} \mathcal{L}(f_\theta(x + \delta),y).
\end{equation}
Here, the model parameters are held fixed, and the search is conducted over the data space. The constraint set will vary depending on the goals of the researcher, the data domain, etc., but a common choice for the constraint is the $\ell_p$-ball of radius $\epsilon$, often with $p=1,2,$ or $\infty$. Additionally, a plethora of approaches to solving this objective under different loss configurations have been proposed, with perhaps the most popular being iterative projected gradient descent (PGD) on the negative cross-entropy loss~\cite{madry2018towards}.

To characterize the adversarial robustness of an AI system, researchers and practitioners alike may be interested in solving Equation~\ref{eq:advpert} under a variety of configurations, which may include swapping out constraints, loss functions, and optimizers. They will need to run this optimization across an entire test set, and may also be interested in re-running this analysis for multiple ML models to compare their robustness. Clearly, assessing adversarial robustness is a complex and computationally-intensive process, requiring tools for solving Equation~\ref{eq:advpert} that are just as composable and scalable as the existing ML tools that solve Equation~\ref{eq:standard}.

Moving beyond the affine perturbation model from Equation~\ref{eq:advpert}, consider the following broad problems that form the core of robust AI, where $\delta$ is now used to represent the parameters of a generalized model for applying transformations to data, $g_\delta(x)$, such as in ~\cite{laidlaw2019functional}:

\begin{description}
\item[Transforming Data (e.g., augmentations, corruptions):]
\begin{equation} \label{eq:transform}
    g_\delta(x)
\end{equation}

\item[Optimizing Transformations of Data:]
\begin{equation} \label{eq:opt1}
    \max\limits_{\delta \in \Delta} \mathcal{L}(f_\theta(g_{\delta}(x)),y)
\end{equation}

\item[Optimizing Transformations of Data Distributions:]
\begin{equation} \label{eq:opt2}
    \max\limits_{\delta \in \Delta} \mathbb{E}_{(x,y)\sim D} [\mathcal{L}(f_\theta(g_\delta(x)),y)]
\end{equation}

\item[Optimizing Models on Transformed Data:]
\begin{equation} \label{eq:opt3}
    \min\limits_{\theta \in \Theta} \mathbb{E}_{(x,y)\sim D} [ \max\limits_{\delta \in \Delta} \mathcal{L}(f_\theta(g_\delta(x)),y) ]  
\end{equation}

\item[Optimizing Models on Transformed Distributions:]
\begin{equation} \label{eq:opt4}
    \min\limits_{\theta \in \Theta} \max\limits_{\delta \in \Delta} \mathbb{E}_{(x,y)\sim D} [\mathcal{L}(f_\theta(g_\delta(x)),y) ]
\end{equation}
\end{description}

Our Responsible AI Toolbox (\texttt{rAI-toolbox}) is designed to support all of the flavors of analysis represented by Equations~\ref{eq:transform} - \ref{eq:opt4}. The solutions to these sophisticated, data-scrutinizing problems, which often depend on the state of the model itself, do not naturally fit into the standard train, test, and deploy stages that are facilitated by popular, high-level ML APIs like
PyTorch~\cite{NEURIPS2019_9015},
PyTorch Lightning~\cite{falcon2020framework}, or \texttt{fastai}~\cite{howard2018fastai}. As a result, much of the tooling that is currently available for assessing and enhancing the robustness in AI systems is research-grade code. Relative to standard APIs, these existing tools lack in composability, scalability, and reliability (i.e., often they are under-tested or have no tests whatsoever). In the remaining three sections, we discuss the design principles behind the \texttt{rAI-toolbox} that address these challenges, along with illustrative examples.

\subsection{Composability via Tight Integration with PyTorch APIs}\label{ss:composability}

A key design principle of the \texttt{rAI-toolbox} is that we adhere strictly to the general APIs specified by the PyTorch machine learning framework. This enables our library's functions and boilerplate code to compose elegantly with the powerful features provided by PyTorch and other PyTorch-based projects, and it ensures that we expose only familiar and well-defined interfaces to our users. In contrast, we have found that attempts by robustness libraries to simultaneously cater to multiple ML frameworks, such as TensorFlow~\cite{tensorflow2015-whitepaper} and JAX~\cite{jax2018github} in addition to PyTorch, inevitably hinders this essential composability.

To highlight the utility of this design principle, consider that the data exploration techniques that are employed in AI robustness analyses (and introduced in Section~\ref{ss:core_problems}) are often substantially more sophisticated than the gradient-based methods that typically support model training. For example, model-dependent line-searches and $\ell_p$-norm projected optimizers are essential tools of the trade in this domain. Despite this sophistication, the \texttt{rAI-toolbox} designs all data exploration workflows around the basic \texttt{torch.optim.Optimizer} API. Thus, users can readily use PyTorch-based ``off-the-shelf" optimizers for robustness analysis and enhancement, or they can implement a custom optimizer towards this end.

The benefits afforded by this design choice are hard to overstate; we can natively accommodate over data domains: line-searches via closures, heterogeneous learning-rate schedulers, higher-order optimization methods, and more. These capabilities are not available in other AI robustness toolkits, which tend to design their APIs in service of particular popular research results. Ultimately, we anticipate that the design of the \texttt{rAI-toolbox} will enable innovative engineering solutions for improving and optimizing AI robustness techniques, as well as the incorporation of these techniques at-scale in MLOps pipelines.

\begin{lstlisting}[frame=single,
%float=tp,
numbers=none,
aboveskip=10pt,
xleftmargin=4pt,
% xleftmargin=0pt,
% xrightmargin=-4pt,
language=Python,
label={lst:opt_loop},
caption={A standard workflow to solve for data perturbations (i.e., Equation \ref{eq:opt1}) using \texttt{rAI-toolbox}. Note that it is by design that this workflow closely matches a standard model-training loop in PyTorch.}]
def solve(perturb, opt, criterion, batch, N):
    data, target = batch
    for i in range(N):
        data_pert = perturb(data)
        objective = criterion(data_pert, target)
      
        # update perturbation parameters
        opt.zero_grad()
        objective.backward()
        opt.step()
\end{lstlisting}

\begin{lstlisting}[frame=single,
% float=tp,
numbers=none,
aboveskip=10pt,
belowskip=10pt,
xleftmargin=4pt,
language=Python,
label={lst:pgd},
caption={Leveraging the solver from Listing \ref{lst:opt_loop} to perform projected gradient descent, in search of minimal data perturbations that ``fool" our model. Some representative results are depicted in Figure \ref{fig:pgd}.}]
from rai_toolbox.perturbations import Affine
from rai_toolbox.optim import L2ProjectedGradient
from torch.nn.functional import cross_entropy

data, target = # load ImageNet data
pert = Affine(data.shape)
opt = L2ProjectedGradient(
    pert.parameters(),
    InnerOpt=SGD,
    epsilon=1.0,
    lr=0.1
)

def criterion(data_perturbed, target):
    logit = model(data_perturbed)
    return -cross_entropy(logit, target)
  
batch = (data, target)
solve(pert, opt, criterion, batch, 10)
\end{lstlisting}

\begin{figure}[ht]
\centering
\includegraphics[width=3.25in]{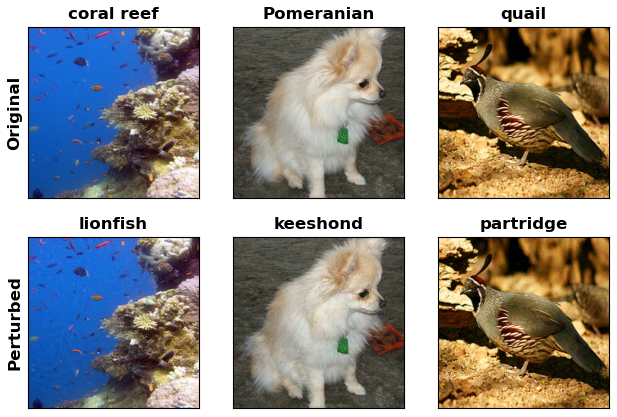}
\caption{Some illustrative results showing original (top row) and adversarially perturbed (bottom row) images and their predicted labels from a standard ImageNet model, generated using the \texttt{rAI-toolbox} via Listings \ref{lst:opt_loop} and \ref{lst:pgd}.}
\label{fig:pgd}
\end{figure}

\begin{lstlisting}[float=tp, frame=single,
numbers=none,
xleftmargin=4pt,
language=Python,
label={lst:xai},
caption={Leveraging the solver from Listing \ref{lst:opt_loop} to perform concept probing of our model; the model is prompted to provide an internally generated, visual representation of a concept. Some representative results are shown in Figure \ref{fig:xai}.}]
from rai_toolbox.perturbations import Affine
from rai_toolbox.optim import L1qFrankWolfe
from torch.nn.functional import cross_entropy

data, target = # load ImageNet data
pert = Affine(data.shape)
opt = L1qFrankWolfe(
    pert.parameters(),
    epsilon=7760,
    lr=1
)

def criterion(data_perturbed, target):
    logit = model(data_perturbed)
    return cross_entropy(logit, target)
  
batch = (data, target)
solve(pert, opt, criterion, batch, 10)
\end{lstlisting}

\begin{figure}[t]
\centering
\includegraphics[width=3.25in]{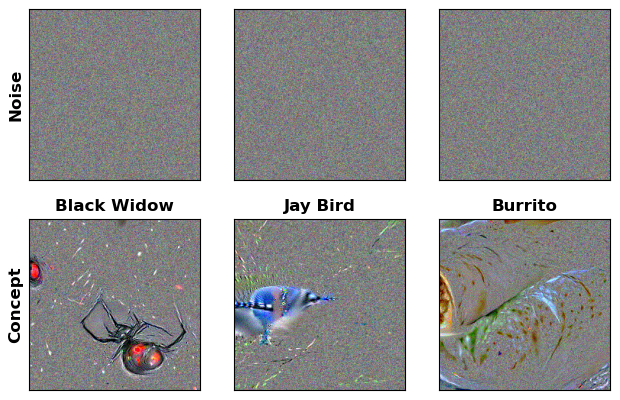}
\caption{Illustrative results of concept probing for a robust ImageNet model, using \texttt{rAI-toolbox} via Listings \ref{lst:opt_loop} and \ref{lst:xai}. Inputs are noisy samples (top) and the optimization is prompted to render representations of a ``Black Widow", ``Jay Bird", and a ``Burrito" (bottom).}
\label{fig:xai}
\end{figure}

To highlight the utility of the design principles discussed in this section, we provide some simple examples of using the \texttt{rAI-toolbox} for two different responsible AI analyses that both involve solving Equation~\ref{eq:opt1}. We use the ImageNet dataset \cite{russakovsky2015imagenet} and standard and robust\footnote{Robust model was trained with $\epsilon=3.0$ and the $\ell_2$ constraint.} models from \cite{madrylab2019robustness} for these examples. Listings \ref{lst:pgd} and \ref{lst:xai} both leverage the same data perturbation solver that is defined in Listing \ref{lst:opt_loop}, yet they perform distinct tasks. The former generates $\ell_2$-constrained adversarial perturbations using PGD (depicted in Figure \ref{fig:pgd}), whereas the latter uses a sparse optimizer to perform concept probing~\cite{roberts2021machine} on a model (depicted in Figure \ref{fig:xai}). The flexibility demonstrated here is a testament to the value of designing the \texttt{rAI-toolbox} around strict adherence to PyTorch's APIs. 

\subsection{Scalability via Distributed Computing Strategies, Hydra, and hydra-zen}\label{ss:scalability}

Scalability is about performing computationally expensive operations in a way that improves productivity and reduces cost of training and testing.  There are two essential properties of scalability that the toolbox provides support for: frameworks supporting PyTorch distributed API to distribute models and data, and scaling experimentation workflows to execute jobs that span multiple models, datasets, and parameters.

First, the \texttt{rAI-toolbox}'s tight integration with PyTorch APIs, such as the \texttt{torch.optim.Optimizer}, provides natural support for frameworks that utilize PyTorch distributed. This is highly beneficial because it helps address integration challenges of techniques that often have large computational costs.  Users can take full advantage of any framework that utilizes the PyTorch distributed interface for data-distributed parallelism, model sharding techniques, and more.  Additionally, integrating robust AI workflows with popular frameworks that support scalability such as PyTorch Lightning or \texttt{fastai} is as simple as integrating traditional training and testing workflows. For example, Listing~\ref{lst:scalable_pgd} demonstrates the simplicity of building a scalable optimization loop for generating adversarial perturbations using PyTorch Lightning's \texttt{LightningLite} API.  Here you will notice the reuse of code provided in Listings~\ref{lst:opt_loop} and~\ref{lst:pgd} with the additional functionality required to setup scalability with \texttt{LightningLite}.  Given PyTorch Lightning's support of multiple distributed strategies, e.g. DeepSpeed~\cite{rasley2020deepspeed}, a user developing with \texttt{rAI-toolbox} will have automatic support for multiple distributed strategies.

\begin{figure}[t]
\vspace*{-0.5cm}
\includegraphics[width=3.2in]{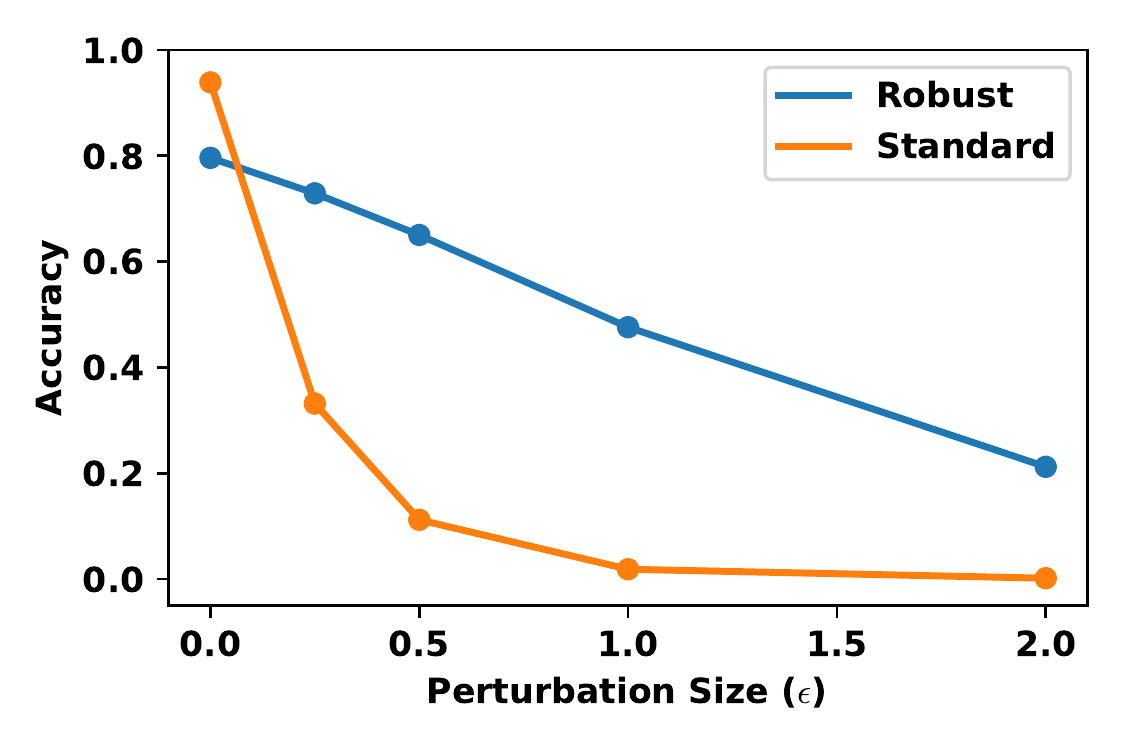}
\centering
\vspace{-0.4cm}
\caption{Robustness curves based on Listing~\ref{lst:scalable_pgd_task}, showing adversarial accuracy on the CIFAR-10 test dataset for a standard and robust model with perturbations of increasing size.}
\label{fig:robustness_curve}
\end{figure}

For scaling experimentation workflows, such as evaluating the robustness of multiple models trained with different techniques, it currently is a challenge to launch multiple jobs in a way that not only distributes each job to reduce the computation time, but also ensures each individual experiment is configurable and repeatable.  This is where we see the power of combining Hydra, \texttt{hydra-zen}, and \texttt{rAI-toolbox}.  As mentioned in Section~\ref{section:hydra-zen}, by configuring experiments with \texttt{hydra-zen}, a user can take full advantage of Hydra's ecosystems of plugins for launching multiple jobs by sweeping over ``swappable'' configurations or ranges on parameter values.  This type of workflow is outlined in Listing~\ref{lst:scalable_pgd_task} to demonstrate how to launch an experiment to compute the performance of a standard and robust model against adversarial perturbations of varying strength (in this case, the size of the $\ell_2$-ball) using PGD.  The outputs of the individual job runs can then be collected and plotted to produce robustness curves, as shown in Figure~\ref{fig:robustness_curve} for a standard and robust\footnote{Robust model was trained with $\epsilon=1.0$ and the $\ell_2$ constraint.} model from \cite{madrylab2019robustness} on the CIFAR-10 dataset \cite{krizhevsky2009learning}. This example demonstrates how the combination of \texttt{hydra-zen} and the \texttt{rAI-toolbox} can enable a complex Responsible AI activity (e.g., evaluating the robustness of multiple ML models) to be conducted at scale and in a traceable manner.

\begin{lstlisting}[float=tp,
frame=single,
numbers=none,
xleftmargin=4pt,
% xleftmargin=0pt,
% xrightmargin=-4pt,
language=Python,
label={lst:scalable_pgd},
caption=An example of scaling \texttt{rAI-toolbox} using PyTorch Lightning's \texttt{LightningLite} to distribute optimization for data perturbations across multiple GPUs. \texttt{LightningLite} handles the distributed computations for the model and data.
]
from pytorch_lightning.lite import LightningLite
from torch.nn.functional import cross_entropy
from rai_toolbox.perturbations import Affine

class Scalable(LightningLite):
    def run(self, model, dataloader, optim, N):
        dataloader = self.setup_dataloaders(dataloader)
        for data, target in dataloader:
            perturb = Affine(data.shape)
            opt = optim(pert.parameters())

            # Combine for distributed
            pmodel = Sequential(perturb, model)
            pmodel, opt = self.setup(pmodel, opt)
            for i in range(N):
                logit = pmodel(data)
                loss = -cross_entropy(logit, target)
                opt.zero_grad()
                self.backward(loss)
                opt.step()
\end{lstlisting}

\begin{lstlisting}[float=*,
numbers=none,
xleftmargin=4pt,
frame=single,
language=Python,
label={lst:scalable_pgd_task},
caption={Configuring and launching an experiment to evaluate performance of a standard and robust model against adversarial perturbations of varying size, ``epsilon''.  \texttt{hydra-zen} is used to build configurations that will be saved out in YAML format for each experiment and launch the jobs which sweep over the models and values of $\epsilon$, while our \texttt{LightningLite} model from Listing~\ref{lst:scalable_pgd} is used to distribute each experiment across 2 GPUs using PyTorch's data distributed parallel.}
]
from hydra_zen import builds, make_config, MISSING, instantiate, launch

Config = make_config(dataloader = builds(...), # config for CIFAR-10 dataloader
  model = builds(...),  # config for either the standard or robust model
  optim = builds(..., epsilon=0.0, zen_partial=True),# Partial config for L2 PGD
  N = 20)

def task_fn(cfg: Config):
    dataloader = instantiate(cfg.dataloader)
    model = instantiate(cfg.model)
    optim = instantiate(cfg.optim)
    scaled_task = Scalable(devices=2, accelerator="gpu", strategy="ddp")
    scaled_task.run(model, dataloader, optim, cfg.N)

# Launch experiment for each combination of model and epsilon
overrides=["model=standard,robust", "optim.epsilon=0.0,0.25,0.5,1.0,2.0"]
launch(Config, task_fn, overrides, multirun=True)
\end{lstlisting}

\subsection{Reliability via Property-Based Testing and Automatic Test-Case Generation}\label{ss:reliability}

A final essential, but oft-overlooked, component to developing analysis tools for AI systems is a comprehensive test suite, which not only exercises the APIs of one's tooling but also checks for the correctness of the software. Given that these tools are being used to diagnose the reliability of already difficult-to-interpret AI systems, their trustworthiness is of manifest importance. That being said, there is a natural barrier to testing such software for correctness under diverse conditions, as there is typically no ``oracle" against which one can test their implementation; it can appear that, in order to test one's implementation of a mathematically sophisticated algorithm, one must obtain an independent implementation to compare against. Rather than indulge in this exercise of tautology, the \texttt{rAI-toolbox} leverages a style of testing known as property-based testing \cite{PBT}, which empowers us to verify that the critical implementation details of our tooling are reliable.

A property-based test (PBT) is designed to check that an expected property of a function holds true under an exceptionally diverse set of inputs to the function. That is, whereas a traditional example-based test (EBT) will check that a function produces exactly the expected output when passed a concrete input, a PBT makes less precise, but more general, assertions about the function. This is an important distinction that enables one to elegantly test complex, scientific software.

For example, suppose that we have implemented in \texttt{rAI-toolbox} a projection function, 
$\Pi: \mathcal{X} \rightarrow \mathcal{X}$,
that maps an input onto a particular sub-domain. An EBT of this function would check that a particular value, $x^\prime$, gets mapped by $\Pi$ to a particular value $y^\prime$, i.e., $\Pi(x^\prime) = y^\prime$. By contrast, a PBT could test that---for \textit{any} input $x \in \mathcal{X}$---$\Pi$ is idempotent: $\Pi(x) = \Pi(\Pi(x))$. Note that the EBT example requires an oracle (e.g., a human performing a calculation by hand) to provide us with $y^\prime$, whereas we can verify the idempotence property of our implementation of $\Pi$ in a purely self-consistent way.
Thus by identifying and testing a sufficiently rich set of properties of a function, which are often metamorphic in nature~\cite{segura2016survey}, we are capable of verifying the correctness of that function over very diverse use cases.

The quality of one's PBTs is largely determined by the diversity and volume of inputs that you can use to exercise a given test. Towards this end, we leverage the test-case generation library Hypothesis~\cite{maciver2019hypothesis,maciver2020test}. Hypothesis is a popular Python library that provides powerful test-case generation strategies that are used to ``drive" PBTs. The primitives that it offers can be composed to create sophisticated descriptions of data, which are then used to adaptively generate diverse test cases in search for falsifiable assertions in a test. Listing \ref{lst:pbt_example} demonstrates a Hypothesis-driven PBT, which generates finite-valued 32-bit float arrays of arbitrary shapes to check for idempotence of a particular NumPy \cite{harris2020array} function. The test-suite for the \texttt{rAI-toolbox} makes extensive use of Hypothesis; it is an essential aspect contributing to the reliability of our code. Ultimately, we recommend that other library authors consider adopting Hypothesis, and property-based testing in general, to substantially improve the quality of software testing methods for AI systems.

\section{Conclusion and Future Work}
The availability of powerful, easy-to-use deep learning libraries greatly accelerated AI research and led to breakthroughs in computer vision, natural language, and  other domains. As the development and deployment of AI-enabled systems becomes more widespread, there is an urgent need for tools that facilitate responsible AI engineering practices. In this work, we presented two software libraries---\texttt{hydra-zen} and the \texttt{rAI-toolbox}---that help researchers and developers create AI systems that are more configurable, reproducible, robust, and explainable. By keeping the core components in the \texttt{rAI-toolbox} strictly aligned with standard PyTorch APIs, the toolbox should remain interoperable with new developments in the PyTorch ecosystem (e.g., related to distributed computing), as well as provide a flexible foundation for additional research and development of responsible AI capabilities. Some near-term directions for \texttt{rAI-toolbox} development include adding domain-agnostic perturbations and support for additional data augmentation-based training approaches, e.g., AugMix~\cite{hendrycks2019augmix}, to enable application of responsible AI techniques for a broad set of domains and to perform more diverse robustness assessments.

\begin{lstlisting}[float=tp,
  numbers=none,
  xleftmargin=4pt,
  %floatplacement=tbp,
  frame=single, 
  language=Python,
  label={lst:pbt_example}, 
  caption={A property-based test, using the Hypothesis test-case generation library, which validates the idempotent property of  the NumPy clip function. By default, running the test will prompt Hypothesis to adaptively generate 100 arrays of various shapes and contents as test cases.}]
import hypothesis as hp
import hypothesis.extra.numpy as st
import numpy as np

def project(x): return np.clip(x, 0, 1)

@hp.given(st.arrays(np.float32,
    st.array_shapes()))
def test_idempotence(x):
    hp.assume(np.isfinite(x).all())
    x1 = project(x)
    x2 = project(x1)
    assert np.allclose(x1, x2)
\end{lstlisting}

\section{Acknowledgements}
DISTRIBUTION STATEMENT A. Approved for public release. Distribution is unlimited. This material is based upon work supported by the Under Secretary of Defense for Research and Engineering under Air Force Contract No. FA8702-15-D-0001. Any opinions, findings, conclusions or recommendations expressed in this material are those of the author(s) and do not necessarily reflect the views of the Under Secretary of Defense for Research and Engineering. © 2021 Massachusetts Institute of Technology. Delivered to the U.S. Government with Unlimited Rights, as defined in DFARS Part 252.227-7013 or 7014 (Feb 2014). Notwithstanding any copyright notice, U.S. Government rights in this work are defined by DFARS 252.227-7013 or DFARS 252.227-7014 as detailed above. Use of this work other than as specifically authorized by the U.S. Government may violate any copyrights that exist in this work.

A portion of this research was sponsored by the United States Air Force Research Laboratory and the United States Air Force Artificial Intelligence Accelerator and was accomplished under Cooperative Agreement Number FA8750-19-2-1000. The views and conclusions contained in this document are those of the authors and should not be interpreted as representing the official policies, either expressed or implied, of the United States Air Force or the U.S. Government. The U.S. Government is authorized to reproduce and distribute reprints for Government purposes notwithstanding any copyright notation herein.

\bibliography{references.bib}
\end{document}